# Surface EMG-Based Inter-Session/Inter-Subject Gesture Recognition by Leveraging Lightweight All-ConvNet and Transfer Learning

Md. Rabiul Islam, *Student Member, IEEE*, Daniel Massicotte, *Senior Member, IEEE*, Philippe Y. Massicotte, and Wei-Ping Zhu, *Senior Member, IEEE*

*Abstract*— Gesture recognition using low-resolution instantaneous high-density surface electromyography (HD-sEMG) images opens up new avenues for the development of more fluid and natural muscle-computer interfaces. However, the data variability between inter-session and inter-subject scenarios presents a great challenge. The existing approaches employed very large and complex deep ConvNet or 2SRNN-based domain adaptation methods to approximate the distribution shift caused by these inter-session and inter-subject data variability. Hence, these methods also require learning over millions of training parameters and a large pre-trained and target domain dataset in both the pre-training and adaptation stages. As a result, it makes high-end resource-bounded and computationally very expensive for deployment in real-time applications. To overcome this problem, we propose a lightweight All-ConvNet+TL model that leverages lightweight All-ConvNet and transfer learning (TL) for the enhancement of *inter-session* and *inter-subject* gesture recognition performance. *The All-ConvNet+TL* model consists solely of convolutional layers, a simple yet efficient framework for learning invariant and discriminative representations to address the distribution shifts caused by inter-session and inter-subject data variability. Experiments on four datasets demonstrate that our proposed methods outperform the most complex existing approaches by a large margin and achieve state-of-the-art results on inter-session and inter-subject scenarios and perform on par or competitively on intra-session gesture recognition. These performance gaps increase even more when a tiny amount (e.g., a single trial) of data is available on the target domain for adaptation. These outstanding experimental results provide evidence that the current state-of-the-art models may be overparameterized for sEMG-based inter-session and inter-subject gesture recognition tasks.

*Index Terms*— Transfer learning, domain adaptation, convolutional neural network, recurrent neural network, feature extraction, muscle-computer interface, surface electromyography, EMG, gesture recognition

## I. Introduction

GESTURE recognition based on surface electromyography (sEMG) signals has been a core technology for developing next-generation muscle-computer interfaces (MCIs). The major application domains of sEMG-based MCIs are non-intrusive control of active prosthesis [1], wheelchairs [2], exoskeletons [3] or neurorehabilitation [4], neuromuscular diagnosis [5] and providing interaction methods for video games [6], [7]. The existing approaches for gesture recognition using sparse multi-channel sEMG sensors and classical machine learning methods – such as linear discriminant analysis (LDA) [8], support vector machines (SVM) [9], hidden Markov model (HMM) [10] – on windowed descriptive and discriminative time-domain, frequency-domain and/or time-frequency-domain sEMG feature space [11], [12-16]. However, these sparse multi-channel sEMG-based methods are not suitable for real-world applications due to their lack of robustness to electrode shift and positioning [17], [18]. In addition, malfunction to any of these sparse-channel electrodes leads to retraining the entire MCI system. Deep learning-based methods have recently been exploited for gesture recognition using sparse multi-channel sEMG [19-20], [31-32], [61], [68] but their performance is still far from optimum [64].

To address this problem, designing and developing more flexible, convenient, and comfortable high-density sEMG (HD-sEMG) based myoelectric sensors and efficient pattern recognition algorithms have been major research directions in recent years [17-18], [21-30], [36]. However, the existing HD-sEMG-based gesture recognition methods [17-18], [28], [30] still rely on the windowed sEMG (e.g., range between 100 ms and 300 ms [33], [34]), which demands finding an optimal window length. The determination of an optimal window length represents a strong trade-off between classification accuracy and controller delay, both of which increase with an increase in window size.

To further address this problem, distinctive patterns within instantaneous sEMG images were first discovered by Geng et al. [21] and M.R. Islam et al. [22] to develop more fluid and natural muscle-computer interfaces (MCIs). The instantaneous values of HD-sEMG signals at each sampling instant were arranged in a 2D grid in accordance with the electrode positioning. Subsequently, this 2D grid was transformed into a grayscale sEMG image. Therefore, an instantaneous sEMG image represents a relative global measure of the physiological

Md. R. Islam, D. Massicotte, and P.Y. Massicotte are with the Laboratory of Signal and System Integration (LSSI), Department of Electrical and Computer Engineering, Université du Québec à Trois-Rivières, Trois-Rivières, QC, G9A 5H7, Canada. (e-mail: md.rabiul.islam@uqtr.ca; daniel.massicotte@uqtr.ca; philippe.massicotte2@uqtr.ca).
W.-P. Zhu is with the Department of Electrical and Computer Engineering, Concordia University, Montreal, H3G 1M8, Canada. (e-mail: weiping@ece.concordia.ca).

This work has been funded by the Natural Sciences and Engineering Research Council of Canada grant, CMC Microsystems, and the Research Chair on Signal and Intelligent high-performance systems.

processes underlying neuromuscular activities at a given time. Consequently, gesture recognition is performed solely with the sEMG images spatially composed from HD-sEMG signals recorded at a specific instant.

Motivated by these prior works, further studies have been conducted on this promising new research direction over the years [23-27], [29], [36]. However, the state-of-the-art methods [21], [23], [24] for sEMG-based gesture recognition either employed very complex deep and wide CNN or an ensemble of these complex networks for improved gesture recognition performance. Despite the significant performance boost achieved by these state-of-the-art models [21], [23], [24], the heavy computational and intensive memory cost hinders deploying them on resource-constrained embedded and mobile devices for real-time applications.

In addition, the sEMG-based gesture recognition problem becomes more challenging in the operational conditions or an *inter-session* scenario, where the trained model is used to recognize muscular activities in a new recording session because sEMG signals are highly subject-specific. The distributions of the sEMG signals vary considerably even between recording sessions of the same subject within the same experimental setup. The acquired sEMG signals in a new recording session (target domain or task) differ from those obtained during the training session (source domain or task) because of electrode shifts, changes in arm posture, and slow time-dependent changes such as fatigue and electrode-skin contact impedance [1][26]. Inter-session is often referred to as inter-subject when the training and test data are acquired from different subjects. Moreover, it is always challenging to force the users to maintain a certain level of muscular contraction force in real-time applications. Therefore, the developed methods must also cope with the distribution shift occurred by this voluntary muscular contraction force level.

To attenuate these distribution shifts between different sEMG recording sessions, the *pre-trained models* have been predominantly adopted by the existing approaches [26], [31], [32], and [57] to reduce the distribution shift by *fine-tuning* the sEMG data recorded in the different session (target domain or task). *Fine-tuning* updates the parameters of the *pre-trained* models to train to newly recorded sEMG data. Generally, the output layer of the pre-trained models is extended with randomly initialized weights. A small learning rate is used to *fine-tune* all the parameters from their original values to minimize the loss on the newly recorded sEMG data. Using appropriate hyper-parameters for training, the resulting *fine-tuned model* often outperforms learning from a randomly initialized network [40].

Generally, this *pre-training* and *fine-tuning* process can be considered a special case of *domain adaptation* when the *source task* and the *target task* are the same or *transfer learning* when the tasks are different. However, for sEMG-based gesture recognition scenarios, we reframed this problem as *transfer learning* when the sEMG data for training and inference are recorded at a different session. Fig. 1 illustrates the conceptual diagram of our proposed transfer-learning methods for sEMG-based gesture recognition.

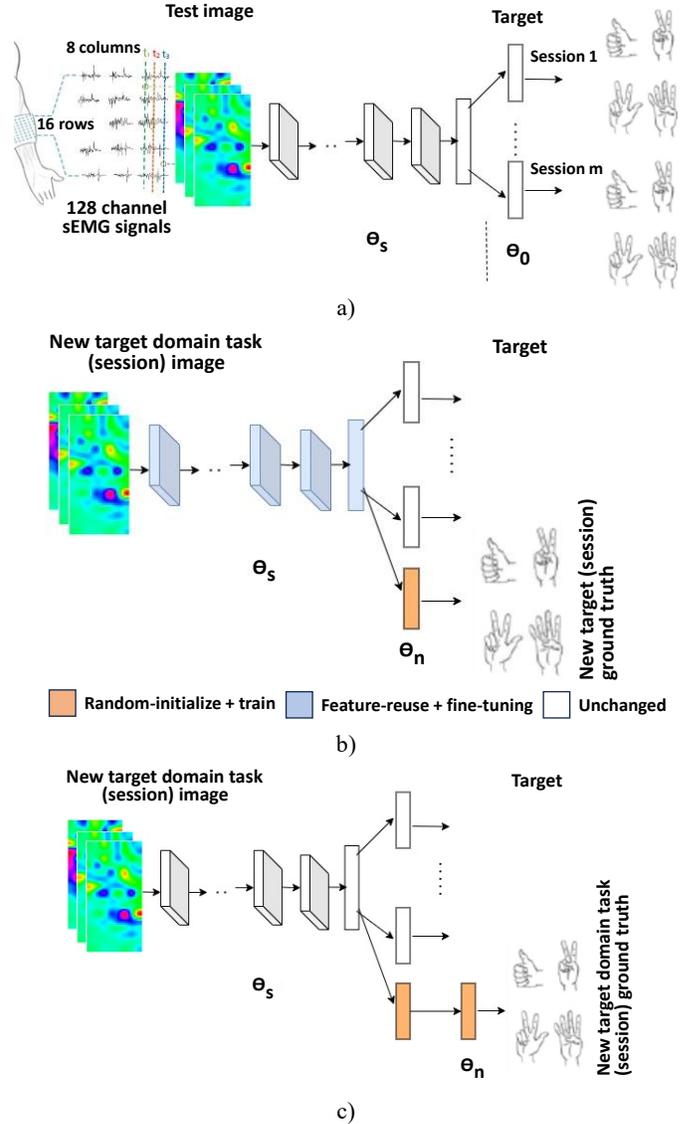

Fig. 1. A general conceptual diagram of the transfer learning method (a) Pre-trained model (b) Fine-tuned model and (c) Feature extraction process. sEMG images and labels used for adaptation are shown.

Transfer learning is typically performed by taking a standard architecture along with its pre-trained weights and then *fine-tuning* the target task. However, the state-of-the-art methods [21], [23], [26], and [61] for sEMG-based gesture recognition employed very large and deep pre-trained models, therefore, containing millions of parameters which are designed to be trained with large-scale labeled sEMG datasets. The requirement of high-end computing resources and large-scale *pre-trained* datasets are also bounded by large and deep network structures [25]. As far as we are aware, there has been no research for sEMG-based gesture recognition studying the effects of transfer learning on the smaller, simpler, and lightweight CNN. This line of investigation is especially crucial in the sEMG-based gesture recognition because the pre-trained model is often deployed in real-time MCI applications such as assistive technology and physical rehabilitation where fine-tuning in the target domain must be conducted in the data-starved condition because of the difficulty of acquiring data from the amputees, elderly peoples, and patients, etc. Also, the large computationally expensive models might significantly impede mobile and on-device applications, where power consumption, data memory, and computational speed are constraints. To investigate the effects of transfer learning for sEMG-based gesture recognition, our research is motivated by the following research questions- *does feature reuse takes place during fine-tuning or transfer learning? And if yes, where exactly is it in the network?*

Investigating feature reuse, we find out that some of the differences from transfer learning are due to the over-parametrization of the state-of-the-art, more complex pre-trained models rather than sophisticated feature reuse. Additionally, we discovered that a simple, lightweight model can outperform the more complex and computationally demanding state-of-the-art network architectures. We isolate where useful feature reuse occurs and outline the implications for more efficient lightweight model exploration.

In this paper, we perform a fine-grained study on fine-tuning and transfer learning for sEMG-based gesture recognition. Our main contributions are:

(1) We introduce All-ConvNet+TL model, which leverages the lightweight All-ConvNet and transfer learning to address the distribution shift in inter-session and inter-subject sEMG-based gesture recognition and evaluate it against the more complex state-of-the-art network architectures. Our proposed method leveraging lightweight All-ConvNet and transfer learning outperforms the state-of-the-art methods by a large margin, both when the data from a single trial or multiple trials are available for *fine-tuning/adaptation*. The outstanding *inter-session* and *inter-subject* gesture recognition performance achieved by the proposed lightweight models raises the question of whether the current state-of-the-art models are overparameterized for the sEMG-based gesture recognition problem.

(2) Using further analysis and weight transfusion experiments, where we partially reuse pre-trained weights, we identify locations where meaningful feature reuse occurs and explore hybrid approaches to transfer learning. These approaches involve using a subset of pre-trained weights and redesigning other parts of the network to make them more lightweight.

(3) We conducted more extensive experiments. A performance evaluation on CapgMyo and its four (4) publicly available HD-sEMG sub-datasets was performed on three different sEMG-based gesture recognition tasks: *intra-session*, *inter-session,* and *inter-subject* scenarios. The results showed that our lightweight models outperformed the more complex state-of-the-art models on various tasks and datasets.

The rest of the paper is structured as follows: Section II reviews current state-of-the-art methods for sEMG-based gesture recognition, Section III presents the proposed transfer learning framework, while Section IV presents the lightweight All-ConvNet model architecture and its design principles. Section V introduces the proposed transfer learning design methodology by leveraging lightweight All-ConvNet (All-ConvNet+TL). Section VI describes the experimental framework, and Section VII demonstrates the state-of-the-art results for inter-session and inter-subject gesture recognition and very competitive results for intra-session gesture recognition, obtained from experiments conducted on CapgMyo and its four (4) sub-datasets. Section VIII highlights the state-of-the-art performance achieved by the proposed All-ConvNet+TL and discusses some important findings. Finally, Section IX provides some conclusive remarks.

## II. RELATED WORK

In this section, we present an overview of current state-of-the art methods for sEMG-based gesture recognition. Many efforts have been devoted to proposing novel deep learning methods to enhance the accuracy of sEMG-based gesture recognition. Geng et al. [21] employed a deep convolutional neural network (CNN or ConvNet) to recognize hand gestures from the sEMG images and showed high recognition accuracy on publicly available benchmark HD-sEMG datasets [15], [17], [26]. M. R. Islam et al. [22] proposed to use Histogram of Oriented Gradients (HoG) as discriminative features and an SVM-based feature classification algorithm for high-density EMG images, achieving accurate classification of 8 gestures [11]. Motivated by [21] and [22], further studies have been conducted in recent years [23-27], [29], [36]. Wei et al. [23] proposed a two-stage convolutional neural network (CNN) with a multi-stream decomposition stage and a fusion stage to learn the correlation between certain muscles and specific gestures. The sEMG image is decomposed into different equally sized image patches based on the layout of the electrode arrays on muscles (e.g., each of eight 8×2 electrode arrays in the CapgMyo database [26] individually produces 8×2 equal-sized sEMG image patches). Then, each of these sEMG image patches is independently and in parallel passed through the convolution layers of a single-stream CNN [21], thereby forming a multi-stream CNN. The learned features from all the single-stream CNNs that form a multi-stream CNN are aggregated and fed to a fusion network for gesture recognition. The reported results showed that multi-stream CNN outperformed single-stream CNN by a small margin. Hu et al. [24] proposed a combined CNN-RNN module to capture both spatial and temporal information of sEMG signals for gesture recognition. The recorded sEMG signals were decomposed into small subsegments using a sliding and overlapping windowing strategy. Each of these sEMG subsegments was converted into an sEMG image and simultaneously passed through a multi-stream CNN built upon [21] for feature extraction. Given the input sequence of the extracted features corresponding to each of the sEMG subsegments, a long short-term memory (LSTM) network was learned individually for gesture recognition. Then, the features learned by each of these LSTMs corresponding to each of these sEMG subsegments were concatenated before being fed to a fully connected and SoftMax layer for gesture recognition. Experimental results indicate that a combined CNN-RNN module outperforms the stand-alone CNN and RNN frameworks, respectively. Encouraged by [38], Chen et al. proposed to use of 3D convolution in the convolutional layers of CNNs for spatial and temporal representation of sEMG images [36]. The 3D convolution is attained by convolving a 3D kernel to the cube formed by stacking multiple adjacent sEMG image frames. The feature maps in the convolution layers of a 3D CNN are connected to multiple adjacent sEMG image frames in the previous layer. Hence, the spatiotemporal information is captured. However, multiple 3D convolutions with distinct kernels are required to apply at the same location of the input to learn representative features, which makes 3D CNN computationally expensive. For example, the exploited 3D CNN in [36] requires learning over >30M (million) parameters when the length of the input cube is set to 10 (i.e., the cube is formed by stacking 10 consecutive sEMG image frames).

However, current state-of-the-art methods [21], [23], [24] employed complex deep and wide CNNs or network ensembles for enhanced gesture recognition performance. For example, Geng et al. [21] exploited a DeepFace [35] like very large and deep CNN (dubbed as GengNet), which requires learning >5.63M (million) training parameters only during *fine-tuning* and *pre-trained* on a very large-scale labeled sEMG training datasets. The complexity of this model grows linearly as the input size is increased due to the use of an unshared weight strategy [27]. Wei et al. [23] used an ensemble of eight (8) single-stream GengNet at the decomposition stage only. Hu et al. [24], used a two-stage ensemble network in which an ensemble of multiple single-stream GengNet was used for spatial feature learning, resulting in multiple sequences of 1-D feature representation. Then, these 1-D feature sequences were passed to an ensemble of LSTM networks before a SoftMax layer recognized the targeted gesture. Hence, deploying these state-of-the-art models [21], [23], and [24] on embedded and mobile devices for real-time applications becomes cumbersome, despite achieving significant performance gains. Therefore, the demand for designing low-cost, lightweight networks is highly increasing for low-end resource-limited embedded and mobile devices.

To overcome these problems, more recently, low-latency and parameter-efficient S-ConvNet [25] and All-ConvNet [27] have been introduced, targeting sEMG-based gesture recognition on low-end devices. S-ConvNet [25] was designed to learn sEMG image representation *from scratch through random initialization.* S-ConvNet consists of a network with convolution layers with the shared kernel, a fully connected layer with a small number of neurons, and an occasional dimensionality reduction performed by stridden CNN, demonstrating very competitive gesture recognition accuracy while needing to be learnt $\approx 1/4th$ *learning parameters using a $\approx 12 \times$ smaller dataset* compared to the more complex and high-end resource-bounded state-of-the-art [21]. A similar CNN architecture to that of S-ConvNet is used by Tam et al. [29] for a fully embedded adaptive real-time sEMG-based gesture recognition. Striving to find a simpler and more efficient lightweight network, in our recent work [27], a new architecture called All-ConvNet was introduced that consists solely of convolutional layers and is designed to be more efficient and less computationally intensive than the existing state-of-the-art models for sEMG-based gesture recognition. Comparing the performance of All-ConvNet to other state-of-the-art models shows that it achieves competitive or state-of-the-art performance on a current benchmark HD-sEMG dataset [26], while being significantly lighter, more efficient, and faster to train and evaluate. *All-ConvNet was designed based on the finding of fact that if the sEMG image area covered by units in the topmost convolutional layer covers a portion of the image large enough to recognize its content (i.e., gesture class we want to recognize).* This leads to predictions of sEMG image classes at different positions which can then simply be averaged over the whole image. Hence, the All-ConvNet becomes robust to translations and geometric distortions, which can be very effective in addressing the electrode shift and positioning problem in sEMG-based gesture recognition.

Moreover, *pre-trained* models have been employed by [26], [31], [32], and [57] to mitigate distribution shifts by *fine-tuning* on the target domain or task for sEMG-based gesture recognition in inter-session and inter-subject scenarios. Currently, Du *et al.* [26] and Ketyko *et al.* [57] present state-of-the-art solutions for sEMG-based gesture recognition in inter-session and inter-subject scenarios. Du *et al.* [26] propose a multi-source extension to the classical adaptive batch normalization (AdaBN) technique [37], combined with their most complex deep and large CNN architecture [21]. They employ AdaBN with the hypothesis that the layer weights contain discriminative knowledge related to different hand gestures, while the statistics of the BatchNorm layer [55] represent discriminative knowledge from different recording sessions in inter-session or inter-subject scenarios [37]. The parameters of the pre-trained model's AdaBN [21] are updated using an unsupervised approach for adaptation in the target domain. However, a drawback of this solution arises when dealing with multiple sources (i.e., multiple subjects), as specific constraints and considerations must be imposed for each source during the pre-training phase of the model [57]. Ketyko *et al.* [57] proposed a 2-Stage recurrent neural networks (2SRNN), where a deep stacked RNN sequence classifier was used for pre-training on the source dataset. Then, the weights of the pre-trained deep-stacked RNN classifier were frozen. At the same time, a fully connected layer without a non-linear activation function was trained in a supervised manner on the target dataset for domain adaptation. More explicitly, the deep-stacked RNN classifier was used as a feature extractor by freezing its weight in the domain adaptation stage. However, ConvNet is computationally more efficient and powerful in extracting discriminative features than RNN, even for classification tasks involving long sequences [58], [59]. Unlike these works, the proposed All-ConvNet+TL model capitalizes the inherent invariant properties of translations and geometric distortions in All-ConvNet and investigates the feasibility of applying transfer learning (TL) on the smaller, simpler, and lightweight All-ConvNet to address the distribution shift and learn invariant, discriminative representations for efficient sEMG-based gesture recognition in inter-session and inter-subject scenarios.

### III. THE PROPOSED TRANSFER LEARNING FRAMEWORK

The proposed transfer learning framework for sEMG-based gesture recognition using instantaneous HD-sEMG images includes the following three major computational components: (i) *a lightweight model development* (ii) *pre-training*, and (iii) *fine-tuning*. A schematic diagram of the proposed transfer learning framework for sEMG-based gesture recognition is shown in Fig. 1. Firstly, we devised a lightweight All-ConvNet model. Secondly, the proposed lightweight All-ConvNet was pre-trained (e.g., Fig. 1a) using a large amount of gesture data acquired by HD-sEMG in a single session or over multiple sessions, which may also involve multiple gestures, trials, and subjects, respectively. Then, the pre-trained model was saved and deployed for subject-specific/personalized classifier development, as sEMG-based wearable devices are usually worn by a single user while executing a target task. Typically, input-side layers that play the role of feature extraction are copied from a pre-trained network and kept frozen or fine-tuned (e.g., Fig. 1b and 1c), in contrast, a top classifier for the target task is randomly initialized and then trained at a slow learning rate. *Fine-tuning* often outperforms training from scratch because the pre-trained model already has a great deal of muscular activity information. Potentially, the pre-trained network could be duplicated and fine-tuned for each new target task [40].

### IV. MODEL DESCRIPTION – THE ALL-CONVOLUTIONAL NEURAL NETWORK (ALL-CONVNET)

The current state-of-the-art methods [21], [23], [26], and [61] for sEMG-based gesture recognition use a large, deep ConvNet architecture similar to the one used in DeepFace [35]. This architecture is designed to be pre-trained on a large-scale labeled HD-sEMG training dataset and requires learning >5.63 million (M) parameters only during fine-tuning. As a result, this large-scale pre-trained model becomes a high-end resource-bounded and computationally very expensive to be practical for real-world MCI applications. Moreover, in their pre-trained ConvNet includes two locally connected (LCN) and three fully connected layers among the other convolutions and a *G*-way fully connected layer. However, the LCN layers used an unshared weight scheme [45] that makes their pre-trained ConvNet even computationally more demanding and very difficult to scale on the target domain task. For example, the learning parameters of [21] increase from $\approx 5.63M$ to $\approx 11M$ with a small enhancement of input HD-sEMG image size from 16×8 to 16×16 due to the use of this unshared weight scheme [27]. Hence, a very large-scale labeled training dataset is required for learning these growing numbers of training parameters [35]. However, the LCN can be beneficial in the application domains where the feature's precise location is dependent on the class labels.

Considering the above-mentioned fact, we investigated the following research questions in [27] : (i) Do we expect the devised networks model to produce a location/translation invariant feature representation? and (ii) Do we need a location-dependent feature representation? Following our findings and building on other recent works that aim to find a simple network architecture, we proposed a lightweight All-ConvNet. This new architecture consists solely of convolutional layers. This simple yet effective framework could learn neuromuscular activity from scratch and yield competitive or even state-of-the-art performance using a ≈12×smaller dataset while reducing the learning parameters from ≈5.63M to only ≈460k than the more complex state-of-the-art for sEMG-based gesture recognition. The All-ConvNet architectural design was adopted based on the following principles and observations:

(i) We hypothesized that different hand gestures produce distinct spatial intensity distributions that remain consistent across multiple trials of the same gesture and distinguishable among different gestures. However, we observed that the spatial intensity distributions for the same gesture are not locally invariant, and the precise feature's location are independent of the class labels. Fig. 2 demonstrates a sequence of HD-sEMG images derived from the same class, along with a correlation heatmap of HD-sEMG distributions (images) sampled equidistantly in time (e.g., each 20 ms) which demonstrates that the distributions are independent of the class labels. CNN alone has a remarkable capability to exploit locally translational invariance features by utilizing local connectivity and weight-sharing strategies [45]. On the other hand, the LCN layer fails to model the relations of parameters in different locations. Hence, the LCN layers are ablated in designing

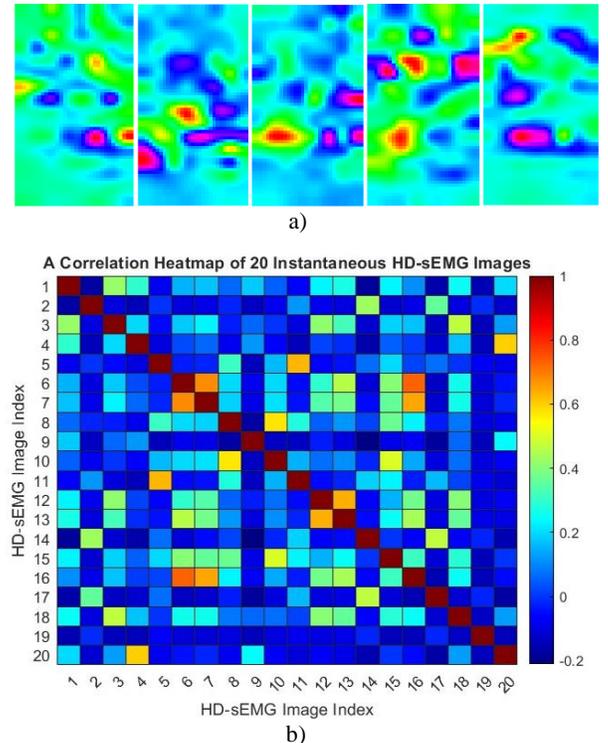

Fig. 2 HD-sEMGs derived from the same muscular activity class (a) and correlation heatmap of HD-sEMG distributions (b) which demonstrates that the distributions are independent to the class labels.

our All-ConvNet models as the location of the features is not dependent on the class labels.

(ii) Inspired by previous work [46], we leverage the fact that if the part of the instantaneous HD-sEMG image is covered by the units in the topmost convolution layers could be large enough to recognize its content (i.e., the gesture class, we want to recognize). Consequently, the fully connected layers can also be replaced by simple 1-by-1 convolutions. This allows us to predict HD-sEMG image classes at different positions, and we can then average these predictions across the entire image. Hence, the proposed All-ConvNet can be very effective in addressing the electrode shift and positioning problem for sEMG-based gesture recognition, where the entire sEMG data stream for a particular gesture may not necessarily be required for recognition. Lin et al. [47], initially introduced this approach, which acts as an additional regularization technique due to the significantly fewer parameters of a 1-by-1 convolution in comparison to a fully connected and LCN layers. Overall, our architecture is thus reduced to consist only of convolutional layers with ELU non-linearities [48], [63] and a global average pooling (GAP) + SoftMax layer to produce predictions over the entire instantaneous HD-sEMG image. A conceptual diagram of our proposed pre-trained All-ConvNet is shown in Fig. 1(a). Table I describes our proposed All-ConvNet architecture. The feature maps learned by the proposed All-ConvNet are presented in Fig. 3.

We train our proposed All-ConvNet for a multi-class sEMG-based gesture recognition task, which involves recognizing a specific muscular activity class using an instantaneous HD-sEMG image. As described in Table I, in the proposed All-ConvNet network, we consider using 1-by-1 convolution at the top to produce 8 or 12 outputs (depending on the number of distinct movements performed). These outputs were then averaged across all positions and fed into a *G*-way SoftMax layer (where *G* is the number of distinct hand gesture classes) which produces a distribution over the class labels. In order to estimate the class probabilities, we use the SoftMax function $\sigma(\cdot)$ with $\hat{y}^{(j)}$ representing the *j*th element of the *G* dimensional output vector of the layer preceding the SoftMax layer, defined as below:

$$\sigma(\hat{y}^{(j)}) = \frac{\exp(\hat{y}^{(j)})}{\sum_G \exp(\hat{y}^{(G)})} \quad (1)$$

The objective of this training is to maximize the probability of the correct gesture class. This is accomplished by minimizing the cross-entropy loss [49] for each training sample. When $y$ represents the true label for a given input, the loss is computed as:

$$L = -\sum_j y^{(j)} \ln(\sigma(\hat{y}^{(j)})) \quad (2)$$

The loss is minimized over the parameters by computing the gradient of $L$ with respect to the parameters. These parameters are then updated using the state-of-the-art Adam (adaptive moment estimation) gradient descent-based optimization algorithm [50]. This algorithm provides fast and reliable learning convergence, unlike the stochastic gradient descent (SGD) optimization algorithm used in state-of-the-art pre-trained networks for gesture recognition using instantaneous HD-sEMG image recognition.

TABLE I THE ALL-CONVNET NETWORK MODEL FOR NEUROMUSCULAR ACTIVITY RECOGNITION.

| All-ConvNet |
|---|
| Input 16×16 Gray-level Image |
| 3 × 3 Conv.64 ELU |
| 3 × 3 Conv.64 ELU |
| 3 × 3 Conv. 64 ELU with stride *r* =2 |
| 3× 3 Conv. 128 ELU |
| 3× 3 Conv. 128 ELU |
| 3× 3 Conv. 128 ELU with stride *r* =2 |
| 1×1 Conv. 128 ELU |
| 1×1 Conv. 8 ELU |
| global averaging over 4×4 spatial dimensions |
| G-way SoftMax |

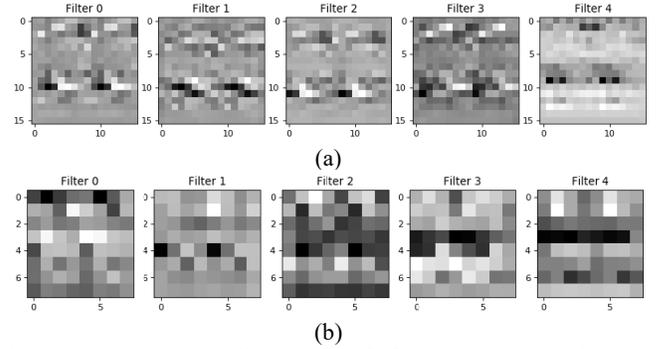

(a)

(b)

Fig. 3. A schematic illustration of feature maps obtained by All-ConvNet before and after dimensionality reduction. (a) Feature maps and b) Feature maps after dimensionality reduction.

Once the network has been trained, an instantaneous HD-sEMG image is recognized as in the gesture class $C$ by simply propagating the input image forward and computing:

$$C = argmax_j(\hat{y}^{(j)}) \quad (3)$$

## V. TRANSFER LEARNING BY LEVERAGING LIGHTWEIGHT ALL-CONVNET (ALL-CONVNET+TL)

In this section, we introduce some notations and definitions used in our transfer learning framework as in [51]. We denote the *source domain* data as $D_s = \{(xs_1, ys_1), ..., (xs_{n_S}, ys_{n_S})\}$, where $xs_i \in X_S$ is the data instance and $ys_i \in Y_S$ is the corresponding class label. In our sEMG-based gesture recognition example, $D_s$ can be a set of sEMG data of different gestures and their corresponding gesture class labels acquired by a single or multiple participants in a designated session. An objective function $f_s(.)$ can be learned using $D_s$ for the source task such that, $T_s = \{Y_s, f_s(\sum_i w_{s_i} X_S + b)\}$. Similarly, we denote the *target domain* data as $D_T = \{(xT_1, yT_1), ..., (xT_{n_T}, yT_{n_T})\}$ and $T_T = \{Y_T, f_T(\sum_i w_{T_i} X_T + b)\}$, where, $xT_i \in X_T$ and $yT_i \in Y_T$ are the sEMG data of different gestures and their corresponding class labels respectively acquired by a distinct subject/participant at a different session than $D_s$. In most cases, the target *domain data* for a distinct participant acquired at another session is much lower quantities than that of a source *domain data, i.e.* $0 \leq n_T \ll n_s$.

Now we define our proposed transfer learning problem as follows– *Given a source domain $D_S$ and a learning task $\mathcal{T}_S$ as well as a target domain $D_T$ and learning task $\mathcal{T}_T$, the transfer learning aims to help improve the learning of the target predictive function $f_T(.)$ in $D_T$ using the knowledge in $D_S$ and $\mathcal{T}_S$, where, $D_S \neq D_T$, and $\mathcal{T}_S = \mathcal{T}_T$. In our sEMG-based gesture recognition problem, the source and target task are the same. However, the data distribution between the source and the target domain might be different i.e., $D_S \neq D_T$ due to factors described in section I.*

To mitigate these distribution shifts on the sEMG-based gesture recognition problem, we apply the transfer learning to our proposed lightweight All-ConvNet [27] and termed it as All-ConvNet+TL. In our setting, All-ConvNet+TL has a set of shared parameters $\theta_s$ (e.g., all the convolutional layers in All-ConvNet) and task-specific parameters for previously learned gesture recognition tasks $\theta_0$ (e.g., the output layer of All-ConvNet for gesture recognition and its corresponding weights), and the task-specific parameters are randomly initialized for new target tasks $\theta_n$ (e.g., gesture recognition in a new session). Considering $\theta_0$ and $\theta_n$ as classifiers that operate on features parameterized by $\theta_s$. Drawing motivation from [40], [65-66], in this work, we adopt the following approaches to learning $\theta_n$ while taking advantage of previously learned $\theta_s$, which is illustrated in Fig. 1:

(i) **Fine-tuning** – involves optimizing $\theta_s$ and $\theta_n$ for the new target task, while keeping $\theta_0$ fixed (as shown in Fig.1b). To prevent large drift in $\theta_s$, a low learning rate is usually used. It is possible to duplicate the original network and fine-tune it for each new target task to create a set of specialized networks.

(ii) **Feature Extraction** – $\theta_s$ and $\theta_0$ remain fixed and unchanged, while the outputs of one or more layers are used as features for the new target task in training $\theta_n$ (as shown in Fig. 1c).

The most popular methodology for transfer learning is to duplicate the pre-trained network (i.e., initialize from pre-trained weights) and fine-tune (train) the entire network for each new target task [62]. However, *fine-tuning* degrades performance on previously learned tasks from the source dataset because the shared parameters change without receiving new guidance for the source-task-specific prediction parameters. In addition, *duplicating* and *fine-tuning* all the parameters of a *pre-trained model* may also require a substantial amount of target task dataset. On the other hand, *feature extraction* usually underperforms on the target dataset because the shared parameters often fail to effectively capture some discriminative information that is crucial for the target task. To address this problem and find out a good trade-off between fine-tuning and feature extraction, we focus on answering the following research questions – *Does feature reuse take place during fine-tuning or transfer learning? And if yes, where exactly is it in the network?* We first conducted a preliminary weight (or feature) transfusion experiment, where we partially reused pre-trained weights to determine and isolate the locations where meaningful feature reuse occurs. We perform this via a weight transfusion experiment by transferring a contiguous set of some of the pre-trained weights, randomly initializing the rest of the network, and training on the target task. We have found out that meaningful feature reuse is restricted to the lowest few layers of the network and is supported by gesture recognition accuracy and convergence speed (see Appendix A for details). Following the results of these weight (or feature) transfusion experiments, the part of the $\theta_s$ (i.e., the first three convolutional layers of All-ConvNet) were frozen and used as a *feature extractor* and only $\theta_s$ in the top convolutional layers were *fine-tuned*. Hence, the proposed network model allows the target task to leverage complex features learned from the source dataset and make these features more discriminative for the target task by *fine-tuning* the top convolutional layers. These transfusion results suggest we propose hybrid and more flexible approaches to transfer learning (see Appendix B).

## VI. EXPERIMENTAL SETUP

We evaluated our proposed approach on CapgMyo[1] dataset [26] for studying and quantifying the effects of transfer learning on the smaller, simpler, and lightweight CNN. The CapgMyo dataset was developed to provide a standard benchmark database (DB) to explore new possibilities for studying and the development of cutting-edge muscle-computer interfaces (MCIs). The CapgMyo dataset includes HD-sEMG data for 128 channels (electrodes) acquired from 23 able-bodied subjects ranging in age from 23 to 26 years, which encompasses the majority of the gestures (finger movements) encountered in activities of daily living (see in Appendix C). The sampling rate is 1000 Hz. It comprised 3 sub-databases as follows:

(a) DB-a: contains 8 isometric and isotonic hand gestures obtained from 18 of the 23 subjects. Each gesture was performed and held for 3 to 10 s.
(b) DB-b: contains the same gesture set as in DB-a but was obtained from 10 of the 23 subjects. Each gesture in DB-b was performed and held for approximately 3 seconds. In addition, every subject in DB-b contributed to two separate recording sessions (DB-b Session 1 and DB-b Session 2), with an inter-recording interval greater than 7 (seven) days. Inevitably, the electrodes of the array were attached at slightly different positions at subsequent recording sessions.
(c) DB-c: contains 12 hand gestures (basic movements of the fingers) obtained from 10 of the 23 subjects. Each gesture in DB-c was performed and held for approximately 3 s as in DB-b.

From the viewpoint of MCI application scenarios, the sEMG-based gesture recognition can be categorized into three (3) scenarios:

A. *intra-session*, in which a classifier is trained on the part of the data recorded from the subjects during one session and evaluated on another part of the data recorded from the same session,
B. *inter-session*, in which a classifier is trained on the data recorded from the subjects in one session and tested on the data recorded in another session, and

---
[1] The dataset is made publicly accessible from the following website: *http://zju-capg.org/research_en_electro_capgmyo.html*).

C. *inter-subject,* when a classifier is trained on the data from a group of subjects and tested on the data from an unseen subject.

All three sub-databases (DB-a, DB-b, and DB-c) were used for *intra-session* performance evaluation. *Inter-session* recognition of hand gestures based on sEMG typically suffers from electrode shift and positioning. Therefore, DB-b was used for *inter-session* performance evaluation. Finally, both DB-b Session 2 and DB-c were used for inter-subject performance evaluation.

For CapgMyo database, first, the power-line interferences were removed from the acquired HD-sEMG signals using a 2nd order Butterworth filter with a band-stop range between 45 and 55 Hz. Then, the HD-sEMG signals were arranged in a 2-D grid according to their electrode positioning at each sampling instant. Afterward, this grid was transformed into an instantaneous sEMG image by linearly converting the values of sEMG signals from $mV$ to color intensity as $[-2.5mV, 2.5mV]$ to [0 255]. As a result, instantaneous grayscale sEMG images with a size of 16 × 8 matrices were obtained. To facilitate GAP, we enhance the input HD-sEMG image size from 16×8 to 16×16 using horizontal mirroring. Unlike [21], this enhancement does not increase the learning parameters in the proposed All-ConvNet.

For *pre-training our proposed original model* All-ConvNet, the following configurations were adopted as in [27], the connection weights for All-ConvNet network architecture were randomly initialized using Xavier initialization scheme [52], [53] and the network was trained using Adam optimization algorithm [50]. The momentum decay and scaling decay were initialized to 0.9 and 0.999, respectively. In contrast to SGD employed in [21], [23], and [26], Adam is an adaptive learning rate algorithm, therefore it requires less tuning of the learning rate hyperparameter. For all our experiments, the learning rate of 0.001 was initialized, and smaller batches of 256 randomly chosen samples from the training dataset were fed to the network during consecutive learning iterations. We set a maximum of 100 epochs for training our All-ConvNet model. However, to prevent overfitting, we applied early stopping [54], which interrupts the training process if no improvements in validation loss are observed for 5 consecutive epochs. Batch normalization [55] was applied after the input and before each non-linearity. To further regularize the network, Dropout [56] was applied to all layers with a probability of 25%. The All-ConvNet model was trained on a workstation with an Intel(R) Xeon(R) CPU E5-2620 v3 @ 2.40GHz processor, 32 GB RAM, and an NVIDIA RTX 2080 Ti GPU. Each epoch was completed in approximately 6 s for a test on intra-session gesture recognition. The average inference time per HD-sEMG sample is ≈0.0929 ms on the above-mentioned computational set up. We have also implemented the state-of-the-art network architecture [21] for a fair comparison with our proposed lightweight sEMG-based gesture recognition algorithm. However, we have adopted the same network initialization method, optimization algorithm, and training paradigm as illustrated in [21].

## VII. EXPERIMENTAL RESULTS

The sEMG-based gesture recognition methods in the literature have usually been investigated in intra-session scenarios [21], [23], [24], [36] and [61]. However, in this work, we evaluated the performance of our proposed sEMG-based gesture recognition algorithm by leveraging lightweight All-ConvNet and transfer learning in inter-session and inter-subject scenarios in addition to intra-session gesture recognition. In the following subsections, we evaluated the performance of our proposed lightweight gesture recognition algorithms. We compared them with the state-of-the-art, more complex methods in the above-mentioned three different scenarios.

### A. Intra-Session Performance Evaluation

In this section, we evaluated the performance of sEMG-based gesture recognition in the intra-session scenario. In this scenario, usually, the data variation comes from the difference between the trials and repetitions of the hand/finger gestures performed by an individual. To mitigate this data variations or distribution time shift caused by the repetitions of the gestures in multiple trials in the same session, the state-of-the-art methods performed pre-training their proposed CNN using half of the training data from all the participated subjects (e.g., 18 in DB-a) in the data collection process. Then, the pre-trained model was fine-tuned using the training data from the target subject for the subject-specific classifier development. The major drawback of this approach [21] is that the same training data used for fine-tuning was also seen during pre-training. However, in [27], we argued that the proposed lightweight All-ConvNet trained from scratch using *random initialization* has the great ability to model these distribution shifts caused by the repetitions of hand gestures across multiple trials within the same session. In that setting, we proposed designing and developing a subject-specific individualized classifier using only the sEMG data available for an individual subject while executing a target task without *pre-training*. For example, in CapgMyo DB-a and DB-b, eight (8) isotonic and isometric hand gestures were performed by an individual subject. Each gesture was also trialed and recorded 10 times with a 1000 Hz sampling rate. Thus, an individual subject generates (8×10×1000 = 80,000) instantaneous sEMG images. In CapgMyo DB-c, an individual performed twelve (12) basic movements of the fingers, and hence it generates (12×10×1000 = 120,000) instantaneous sEMG images. For performance evaluation of the proposed subject-specific lightweight All-ConvNet, a leave-one-trial-out cross-validation was performed, in which each of the 10 trials was used in turn as the test set, and the proposed lightweight All-ConvNet was trained and validated using the remaining 9 trials. This entire paradigm of training and testing process is illustrated in Fig. 1a, which shows that only the trained model (without any feature reuse from the pre-trained model) is used for gesture recognition. It is noteworthy that, in [27], we conducted experiments only on the CapgMyo DB-a and reported and compared the results with the state-of-the-art for sEMG-based gesture recognition because the maximum number of subjects (18) participated in DB-a. However, in this work, we extended our experiments on the CapgMyo DB-b and DB-c, respectively. Table II presents the gesture recognition results for the proposed lightweight All-ConvNet and compares them with the state-of-the-art methods.

TABLE II. THE AVERAGE RECOGNITION ACCURACIES (%) OF 8 HAND GESTURES FOR CAPGMYO DB-A AND DB-B FOR 18 AND 10 DIFFERENT SUBJECTS RESPECTIVELY AND 12 GESTURES FOR 10 DIFFERENT SUBJECTS IN DB-C. THE NUMBERS ARE MAJORITY VOTED RESULTS USING 160 MS WINDOW (I.E., 160 FRAMES). PER-FRAME ACCURACIES ARE SHOWN IN PARENTHESIS.

| Model | S-ConvNet [25] | W.Geng et al., [21] | All-ConvNet (proposed) |
|---|---|---|---|
| CapgMyo DB-a | **98.36 (87.95)** | 98.48 (86.92) | 98.02 (86.73) |
| CapgMyo DB-b Session 1 | **97.87 (83.57)** | 97.04 (81.26) | 97.52 (81.95) |
| CapgMyo DB-b Session 2 | **97.05 (84.73)** | 96.26 (83.21) | 96.80 (83.36) |
| CapgMyo DB-c | **95.80 (81.63)** | 96.36 (82.23) | 95.76 (80.91) |
| #Learning Parameters | $\approx 2.09\ M$ | $\approx 5.63\ M$ | $\approx \mathbf{0.46\ M}$ |
| Avg-run time (s) | 191.29 | 804.66 | 224.33 |

As can be seen in Table II, the proposed lightweight All-ConvNet (with around 0.46 million learning parameters) consists of a stack of 3×3 convolutional layers with occasional subsampling by a stride of 2. It is trained from random initialization and outperformed the state-of-the-art, more complex GengNet [21], [23], [24], [26] and [61] on the CapgMyo DB-b Session 1 and Session 2 datasets, respectively, and performs comparably to the S-ConvNet [25]. Additionally, the lightweight All-ConvNet performs very competitively or on par with the GengNet [21] and S-ConvNet [25] on the CapgMyo DB-a and CapgMyo DB-c datasets, respectively.

Fig. 4 (a)-(d) presents the sEMG-based instantaneous (or per-frame) gesture recognition accuracies and their statistical significance obtained through leave-one-trial-out cross-validation for ten different test trials for each of the participating subjects in CapgMyo DB-a, DB-b, and DB-c, respectively. The highest instantaneous (or per-frame) gesture recognition accuracies were 86.73% for DB-a, 81.95% and 83.36% for DB-b (Session 1 and Session 2, respectively), and 80.91% for DB-c. Which were obtained with the proposed lightweight All-ConvNet. The high per-frame gesture recognition accuracies and low standard deviation over multiple test trials and subjects in each of the four HD-sEMG datasets mentioned above reflect the high stability of the proposed lightweight All-ConvNet.

In addition, based on a simple majority voting algorithm, we have obtained very good gesture recognition accuracies. Fig. 5 (a)-(d) presents gesture recognition accuracy with different voting windows using lightweight All-ConvNet. The average gesture recognition accuracy of 94.56% and 95.99% were achieved by a simple majority voting with 32 and 64 instantaneous images (or frames) for the above four (4) HD-sEMG datasets.

The higher gesture recognition accuracies of 98.02%, 97.52%, 96.80%, and 95.76% (as shown in Table II and Fig. 5) can be obtained by the proposed lightweight All-ConvNet and a simple majority voting over the recognition result of 160 frames for DB-a, DB-b (Session 1 and Session 2) and DB-c, respectively. These outstanding results confirm that the proposed lightweight All-ConvNet is highly effective for learning all the invariances for low-resolution instantaneous HD-sEMG image recognition and hence seem to be enough to address the problem of employing high-end resource-bounded fine-tuned pre-trained networks for low-resolution instantaneous HD-sEMG image recognition.

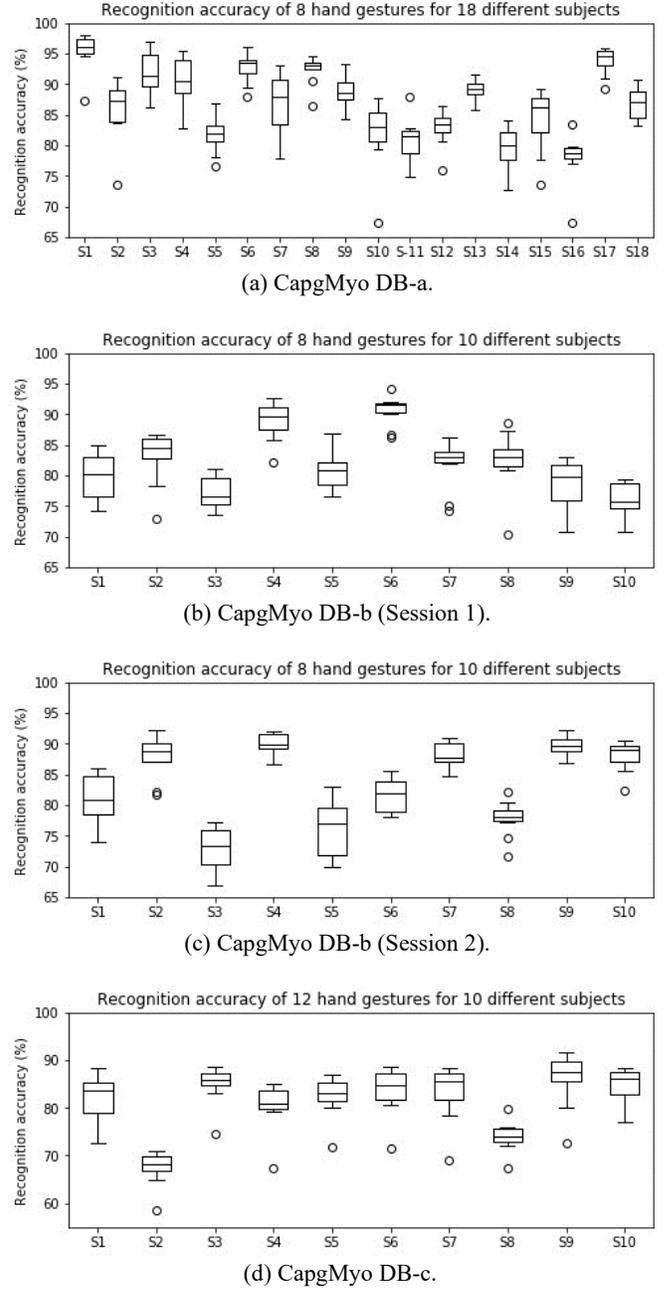

(a) CapgMyo DB-a.

(b) CapgMyo DB-b (Session 1).

(c) CapgMyo DB-b (Session 2).

(d) CapgMyo DB-c.

Fig 4 The per-frame gesture recognition accuracy with our proposed lightweight All-ConvNet, a) the recognition accuracy of 8 hand gestures for 18 different subjects on CapgMyo DB-a, b) and c) The gesture recognition accuracy of 8 hand gestures for 10 different subjects on CapgMyo DB-b (Session 1) and DB-b (Session 2), respectively, and d) the gesture recognition accuracy of 12 hand gestures for 10 different subjects on CapgMyo DB-c.

Table II also includes average run-time for training, validation and inference for an intra-subject test. For a fair run-time comparison, each of the compared models was trained for 100 epochs on the same size of the input HD-sEMG image and early stopping [56] was applied while training all the compared models. The proposed lightweight All-ConvNet exhibits superior run-time performance compared to the state-of-the-art methods.

*B. Inter-Session Performance Evaluation*

In this section, we evaluated the performance of sEMG-based gesture recognition in the inter-session scenario. In this scenario, there is still the intra-session variability discussed in the previous section, in addition to the extent of data variability, which comes from the differences between the recording sessions. The sensor placement may have some spatial shifts and/or rotations at each recording session. These differences in sensor placement and/or rotations may cause spatial shifts in the distributions of the sEMG sensor data. To address this spatial shift problem, currently [26] and [57] provide a state-of-the-art solution in the CapgMyo dataset. Du et al. [26] proposed a multi-source extension to classical AdaBN [37] for domain adaptation. However, when dealing with multiple sources (i.e., multiple subjects), specific constraints and considerations must be imposed for each source during the model's pre-training phase [57]. Ketyko et al. [57] introduced a 2-Stage recurrent neural network (2SRNN) involving pre-training a deep stacked RNN sequence classifier on the source dataset, freezing its weights, and simultaneously training a supervised fully connected layer without a non-linear activation function on the target dataset for domain adaptation. However, ConvNet is more powerful at extracting discriminative features than RNN, even for classification tasks of long sequences [58], [59].

In addition, it is noteworthy that the domain adaptation was conducted in unsupervised and semi-supervised settings [26]. However, very low gesture recognition accuracies were reported in [26] in both inter-session and inter-subject scenarios. On the other hand, [57] performed domain adaptation in supervised settings and demonstrated state-of-the-art results on the CapgMyo dataset. Therefore, for a fair comparison with the state-of-the-art, we performed domain adaptation in a supervised manner in all the compared methods. Moreover, it might be an interesting question why we chose to compare the performance of our proposed lightweight All-ConvNet+TL with the CNN models, proposed in [21] and [26]. To the best of our knowledge, the base CNN models proposed in [21] and [26] were also adapted in [23], [24], and [61], respectively, and reported state-of-the-art results on various sEMG-based gesture recognition tasks and datasets.

Experiments conducted on inter-session and inter-subject settings; we have shown that our proposed lightweight All-ConvNet+TL leveraging transfer learning (illustrated in Section V) outperformed these above-mentioned state-of-the-art solutions. We evaluated inter-session gesture recognition for CapgMyo DBb, in which the model was trained using data recorded from the first session and evaluated using data recorded from the second session. It is worth mentioning that without transfer learning or domain adaptation, the state-of-the-art models, as well as our proposed models achieved less than or approximately 50% average gesture recognition accuracy on CapgMyo datasets in both inter-session and inter-subject scenarios. This level of recognition accuracy is not enough for a usable system (defined as <10% error [60]). Therefore, domain adaptation or transfer learning must be introduced to these (inter-session and inter-subject) settings for acceptable performance. However, the most significant question is how much training data is required for adaptation on the target domain to obtain a stable gesture recognition accuracy. To address this question, we limited the available training data to 20% (T1), 40% (T2), 60% (T3), 80% (T4), and 100% (T5) of the total 5 trials used for domain adaptation (the remaining 5 trials are kept for validation). For fair comparison and complying with the state-of-the-art, we ran our domain adaptation for 100 epochs. Table III presents the *inter-session* average gesture recognition accuracies (%) of 8 hand gestures for 10 different subjects respectively for CapgMyo DB-b and compared with the state-of-the-art methods.

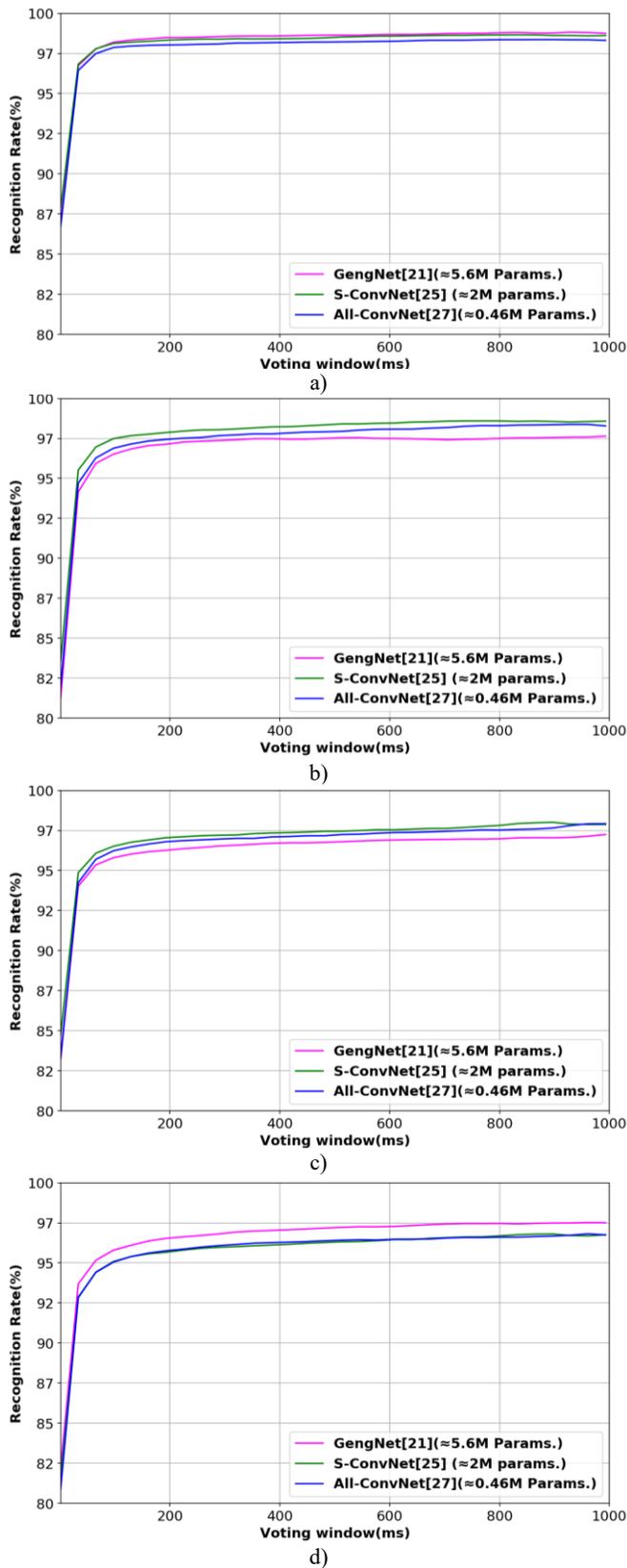

Fig 5 Surface EMG gesture recognition accuracy with different voting windows using the proposed lightweight All-ConvNet and compared with the state-of-the-art methods: a) the recognition accuracy of 8 hand gestures for 18 different subjects on CapgMyo DB-a, and the gesture recognition accuracy of 8 hand gestures for 10 different subjects on CapgMyo for b) DB-b Session 1 and c) DB-b Session 2, and d) the recognition accuracy of 12 hand gestures for 10 different subjects on DB-c.

Our proposed lightweight All-ConvNet+TL leverages transfer learning to enhance inter-session gesture recognition, achieving an 11.11% improvement compared to 2SRNN [57] and a 6.43% improvement compared to GengNet [21][26] when all available 5 trials are used for adaptation (as shown in Table III, column-T5). We also compared our proposed lightweight All-ConvNet+TL with the state-of-the-art GengNet [21][26] in a data-starved condition. The proposed lightweight All-ConvNet+TL shows even more significant improvement over the state-of-the-art when a limited number of trials are available for adaptation, as seen in Table III, Columns- T1, T2, T3, and T4, respectively. For example, the proposed lightweight All-ConvNet+TL achieved a 7.94% improvement over GengNet [21][26] when only 20% of the data (i.e., 1 trial) was available for adaptation (Table III, Column- T1).

### C. Inter-Subject Performance Evaluation

In this section, we evaluated the performance of sEMG-based gesture recognition in the inter-subject scenario. In this scenario, the data variability comes from the variation in muscle physiology between different subjects. In this experiment, we evaluated the inter-subject recognition of 8 gestures using the second recording session of CapgMyo DB-b and the recognition of 12 gestures using CapgMyo DB-c. We performed a leave-one-subject-out cross-validation, in which each of the subjects was used in turn as the test subject, and a lightweight All-ConvNet was pre-trained using the data of the remaining subjects. Then, this pre-trained All-ConvNet model was deployed, and adaptation was made on the data from the odd numbers of trials of the test subjects by leveraging transfer learning or domain adaptation. Finally, the adapted model was evaluated and tested using the data from the even number of trials of the test subject. We limited the available training data to 20%, 40%, 60%, 80%, and 100% of the total 5 trials used for domain adaptation (the remaining 5 trials are kept for validation). Table IV presents the average recognition accuracies (%) of 8 and 12 hand gestures for CapgMyo DB-b and DB-c for 10 subjects, respectively.

As can be seen from Table IV, our proposed lightweight All-ConvNet+TL, by leveraging transfer learning, outperformed the state-of-the-art methods in the inter-subject scenario on both CapgMyo DB-b and CapgMyo DB-c datasets, respectively. Our proposed lightweight All-ConvNet+TL demonstrates an improvement of 5.04% and 6.17% compared to 2SRNN [57], and 3.58% and 1.85% compared to GengNet [21][26] on CapgMyo DB-b and CapgMyo DB-c datasets, respectively when all available 5 trials are used for adaptation (as shown in

TABLE III. INTER-SESSION GESTURE RECOGNITION ACCURACIES ON CAPGMYO DB-B. THE AVERAGE RECOGNITION ACCURACIES (%) OF 8 HAND GESTURES FOR 10 DIFFERENT SUBJECTS RESPECTIVELY. THE NUMBERS ARE THE MAJORITY VOTED RESULTS USING 150 MS WINDOW (I.E., 150 FRAMES).

| Methods | Number of available trials for adaptation | | | | |
|---|---|---|---|---|---|
| | T1 | T2 | T3 | T4 | T5 |
| Du et. al. [21][26] | 67.97 | 81.77 | 86.02 | 88.10 | 88.48 |
| 2SRNN [57] | - | - | - | - | 83.80 |
| All-ConvNet+TL (Proposed) | **75.91** | **89.61** | **92.74** | **93.46** | **94.91** |

TABLE IV. INTER-SUBJECT GESTURE RECOGNITION ACCURACIES. THE AVERAGE RECOGNITION ACCURACIES (%) OF 8 HAND GESTURES FOR CAPGMYO DB-B AND 12 HAND GESTURES FOR CAPGMYO DB-C FOR 10 DIFFERENT SUBJECTS RESPECTIVELY. THE NUMBERS ARE THE MAJORITY VOTED RESULTS USING 150 MS WINDOW (I.E., 150 FRAMES).

| Methods | CapgMyo DB-b | | | | |
|---|---|---|---|---|---|
| | Number of available trials for adaptation | | | | |
| | T1 | T2 | T3 | T4 | T5 |
| Du et. al. [21],[26] | 71.81 | 86.52 | 88.66 | 90.32 | 91.36 |
| 2SRNN [57] | - | - | - | - | 89.90 |
| All-ConvNet+TL (Proposed) | **75.34** | **89.42** | **92.09** | **93.83** | **94.94** |
| | CapgMyo DB-c | | | | |
| Du et. al. [21],[26] | 57.40 | 75.98 | 82.51 | 85.98 | 88.02 |
| 2SRNN [57] | - | - | - | - | 85.40 |
| All-ConvNet+TL (Proposed) | **58.47** | **78.89** | **86.02** | **89.99** | **91.57** |

Table IV, column-T5 for both CapgMyo DB-b and CapgMyo DB-c).

Similar to the inter-session scenario, we also compared our proposed lightweight All-ConvNet+TL in the inter-subject scenario with the state-of-the-art GengNet [21], [26] in a data-starved condition. The proposed lightweight All-ConvNet+TL exhibits improvement over the state-of-the-art on CapgMyo DB-b and CapgMyo DB-c datasets when a limited number of trials are available for adaptation, as observed in Table IV, specifically in Columns T1, T2, T3, and T4, respectively. For example, when only 20% of the data (i.e., 1 trial) was available for adaptation, the proposed lightweight All-ConvNet+TL achieved a 3.53% and 1.07% improvement over GengNet [21], [26] on CapgMyo DB-b and CapgMyo DB-c, respectively (Table IV, Column- T1).

We summarise the inter-session and inter-subject improvement results in Table V over the state-of-the-art methods. As indicated there, the performance of the proposed lightweight All-ConvNet+TL is superior in all cases. The improvement achieved by the lightweight All-ConvNet+TL leveraging transfer learning in inter-session and inter-subject scenarios, exceeds those obtained through alternative state-of-the-art domain adaptation approaches.

Finally, we evaluate the performance of our proposed lightweight All-ConvNet+TL while freezing its maximum number of layers and use them as a feature extractor, and only the top convolutions layers are fine-tuned in the adaptation stage for inter-session and inter-subject gesture recognition. More explicitly, the first six (6) convolutional layers of the lightweight All-ConvNet+TL were frozen and used as a *feature extractor*. Only the top two convolutional layers with a few parameters were fine-tuned in the adaptation stage. Therefore, these experiments can be considered as a full feature extraction setting. The performance of these full feature extraction settings was compared with the more complex computationally expensive 2SRNN [57] method. A deep-stacked RNN classifier was also used as a feature extractor by freezing its weight in the domain adaptation stage. Table VI presents the inter-session and inter-subject average gesture recognition accuracies (%) of 8 and 12 hand gestures for CapgMyo DB-b and DB-c for 10 subjects, respectively. As can be seen from Table VI, our proposed lightweight All-ConvNet+TL clearly outperforms the 2SRNN [57] in both *inter-session* and *inter-subject* gesture recognition accuracy. These experimental results indicate that the proposed lightweight All-ConvNet+TL is very effective for discriminative feature extraction for improved gesture recognition in both inter-session and inter-subject scenarios.

TABLE V. INTER-SESSION AND INTER-SUBJECT IMPROVEMENT (%) RESULTS OBTAINED BY THE PROPOSED LIGHTWEIGHT ALL-CONVNET+TL LEVERAGING TRANSFER LEARNING.

| Methods | Inter-session improvement | Inter-subject improvement | |
|---|---|---|---|
| | DB-b | DB-b | DB-c |
| Du et. al. [21][26] | 6.43 | 3.58 | 3.55 |
| 2SRNN [57] | **11.11** | **5.04** | **6.17** |

TABLE VI. INTER-SESSION AND INTER-SUBJECT GESTURE RECOGNITION ACCURACIES (%) UNDER FULL FEATURE EXTRACTION SETTING.

| Methods | Inter-session | Inter-subject | |
|---|---|---|---|
| | DB-b | DB-b | DB-c |
| 2SRNN [57] | 83.80 | 89.90 | 85.40 |
| All-ConvNet+TL (Proposed) | **91.93** | **91.56** | **85.56** |

## VIII. DISCUSSION

We address the problem of distribution shifts by adapting a lightweight model to new target domain tasks using a limited amount of data for sEMG-based inter-session and inter-subject gesture recognition. We propose All-ConvNet+TL leveraging lightweight All-ConvNet and transfer learning, which can be seen as a hybrid of feature extraction and fine-tuning, learning parameters that are discriminative for the new target task. We show the effectiveness of our method by conducting extensive experiments on CapgMyo and its four (4) publicly available HD-sEMG sub-datasets for three (3) different sEMG-based gesture recognition tasks, including *intra-session*, *inter-session,* and *inter-subject* scenarios. The results indicate that our proposed lightweight All-ConvNet and All-ConvNet+TL models outperform the more complex state-of-the-art models on various tasks and datasets.

In *intra-session* scenarios, the proposed lightweight All-ConvNet (size of only 0.46 M learning parameters), which consists of a network using nothing, but convolutions and subsampling outperformed the most complex state-of-the-art GengNet [21], [26] (size of 5.6M parameters) on CapgMyo DB-b (Session 1 and Session 2) dataset, respectively and performed on par with or very competitively on CapgMyo DB-a and CapgMyo DB-c, respectively. The high *intra-session* gesture recognition accuracies of 98.02%, 97.52%, 96.80%, and 95.76% were obtained by the proposed lightweight All-ConvNet using a simple majority voting over the recognition result of 160 instantaneous images (or frames) for DB-a, DB-b (Session 1 and Session 2) and DB-c, respectively. For gesture recognition in *inter-session* and *inter-subject* scenarios, we apply transfer learning to our proposed lightweight All-ConvNet. Our proposed method All-ConvNet+TL leveraging the lightweight All-ConvNet and transfer learning outperforms the current state-of-the-art methods *by a large margin*, both when the data from *single trials* or *multiple trials* are available for fine-tuning and adaptation.

We achieved state-of-the-art performance for inter-session and inter-subject scenarios. The *inter-session* gesture recognition accuracy reached 94.1% on CapgMyo DB-b, which is approximately 11.11% and 6.43% higher than the current state-of-the-art [57] and [21][26], respectively.

In addition, the *inter-subject* gesture recognition accuracy reached 94.94% and 91.57% on CapgMyo DB-b and DB-c, respectively, which is 5.04% and 6.17% higher than [57] and 3.58% and 3.55% higher than the [21], [26] respectively. Moreover, the proposed lightweight models achieved state-of-art performance under full feature extraction settings in both inter-session and inter-subject scenarios.

These outstanding *state-of-the-art* inter-session and inter-subject gesture recognition performance achieved by the proposed lightweight All-ConvNet+TL models by leveraging transfer learning validates that the proposed method is highly effective in learning invariant and discriminative representations to overcome the distribution shift caused by inter-session and inter-subject data variability. This potentially indicates that the current state-of-the-art models are overparameterized for the sEMG-based gesture recognition problem.

Furthermore, the current most complex state-of-the-art models [21], [26], [57] are computationally expensive and require a huge memory space to store a massive number of parameters. Therefore, these models are usually unsuitable for deploying low-end, resource-constrained embedded and mobile devices for real-time MCI applications. Thanks to the proposed parameter-efficient All-ConvNet and All-ConvNet+TL, our model is much smaller and lightweight than these current state-of-the-art methods for sEMG-based gesture recognition.

Finally, the new experimental evidence of our proposed method about various sEMG-based gesture recognition tasks and its role will shed light on potential future directions for the community to move forward for more efficient lightweight model exploration.

## IX. CONCLUSION

For real-time Muscle-Computer Interfaces, the sEMG-based gesture recognition must address the *inter-session* and *inter-subject* distribution shifts. To address and overcome these distribution shifts, we investigate the effects of transfer learning and feature reuse on our proposed lightweight All-ConvNet. We discovered that the proposed lightweight All-ConvNet+TL, which leverages transfer learning in the *inter-session* and *inter-subject* scenarios outperforms the most complex state-of-the-art domain adaptation methods by a large margin, both when the data from single trials or multiple trials are available for *adaptation*. The state-of-the-art performance proved that the proposed lightweight All-ConvNet+TL model is highly effective in learning invariant and discriminative representations for addressing distribution shifts in sEMG-based inter-session and inter-subject gesture recognition. This raises the question and provides evidence of overparameterization of the most complex current state-of-the-art models for sEMG-based gesture recognition tasks. We also find that significant feature reuse concentrated in lower layers and explored more flexible and hybrid transfer approaches, which retain transfer benefits and create new possibilities. In future work, we plan to deploy our proposed lightweight All-ConvNet and All-ConvNet+TL model for sEMG-based real-time adaptive and intuitive control of an active prosthesis.

**Appendix** to "Surface EMG-Based Inter-Session/Inter-Subject Gesture Recognition by Leveraging Lightweight All-ConvNet and Transfer Learning."

*A. Weight (or Feature) Transfusion Experiments*

In this section, we investigate to identify locations where exactly in the network meaningful feature reuse takes place during transfer learning by conducting a weight (or feature) transfusion experiment. We initialize our proposed lightweight All-ConvNet+TL with a contiguous subset of the layers using pre-trained weights (weight transfusion), and the rest of the network randomly, and train on the target inter-session gesture recognition task. More explicitly, we initialize only up to layer L with pretrained lightweight All-ConvNet+TL weights, and layer L+1 onwards randomly; then train only layers L+1 onwards. Since, the weight transfusion process uses pre-trained weights, it can accelerate the training during fine-tuning of a network on the target task. Therefore, the learning speed was measured in terms of gesture recognition performance on various training epochs. Table VII presents the inter-session gesture recognition accuracy of a subject against various training epochs for different number of transfused weights. We show the learning speed and gesture recognition accuracy when transfusing from Conv1 (L-7, one layer) up to Conv8 (i.e., layer L-7 to layers L-full transfer). From the weight transfusion results, our proposed lightweight All-ConvNet+TL model perform quite stably over the different number of transfused weights. However, we observed that reusing the lowest layers (transfusing weights) leads to the greatest gain in learning speed and gesture recognition accuracy. For example, transfusing weights from layer L-7 (Conv1) up to layer L-5 (Conv3), we achieve ≈ 98% recognition accuracy after just 8 (eight) training epochs.

*B. Lightweight All-ConvNet Network Trimming*

These weight transfusion results (Appendix A) motivate us to explore hybrid approaches to transfer learning, thereby, we introduce network trimming which further optimizes the proposed lightweight All-ConvNet+TL by pruning the weights of the network. We consider reusing pre-trained weights up to Conv3 (i.e., weights of layers L-7 to layers L-5 showed in Table VII) and the weights of the top of the lightweight All-ConvNet (i.e., from layers Conv4 (L-4) to Conv7 (L-1)) was pruned by halves to be even more lightweight and initializing these layers randomly. Finally, this new Lightweight All-ConvNet-Slim model was trained or fine-tuned on the target inter-session gesture recognition task. Table VIII presents the inter-session gesture recognition accuracy of a subject against various training epochs, which compares the performance of Lightweight All-ConvNet+TL vs Lightweight All-ConvNet-Slim model. The experimental results demonstrates that the lightweight All-ConvNet-Slim model can maintain the same or achieve higher performance with much smaller number of parameters. These results with variants of Lightweight All-ConvNet+TL model also highlight many new, rich and flexible ways to use transfer learning. The preprint version of this paper has been made publicly available in [67].

TABLE VII. LEARNING (OR CONVERGENCE) SPEED USING VARIOUS TRAINING EPOCHS. TABLE SHOWS INTER-SESSION GESTURE RECOGNITION ACCURACIES (%) ON TEST SET. THE NUMBERS ARE MAJORITY VOTED RESULTS USING 150 MS WINDOW (I.E., 150 FRAMES). PER-FRAME ACCURACIES ARE SHOWN IN PARENTHESIS.

| Weight transfusion (up to layers) | Training epochs | | | | | |
|---|---|---|---|---|---|---|
| | 8 | 16 | 32 | 46 | 64 | 100 |
| Full Transfer (L) | 70.90 (64.56) | 81.74 (67.84) | 83.20 (68.35) | 83.08 (68.33) | 83.21 (68.47) | **83.60** (68.52) |
| L-1 | 87.42 (72.28) | 88.21 (73.53) | 90.14 (74.43) | 90.01 (74.55) | 89.85 (74.94) | **90.39** (75.13) |
| L-2 | 90.24 (76.35) | 93.60 (78.17) | 93.94 (79.62) | 94.22 (80.08) | **94.50** (80.47) | 94.18 (81.36) |
| L-3 | 95.01 (79.48) | 95.96 (81.53) | 96.42 (83.23) | 96.71 (83.22) | 96.99 (83.97) | **98.28** (84.67) |
| L-4 | 96.10 (81.87) | 97.71 (82.59) | 98.21 (85.10) | 97.92 (86.17) | 97.96 (86.37) | **98.59** (87.06) |
| L-5 | 97.96 (83.14) | 98.40 (84.888) | 99.12 (87.00) | 99.12 (86.99) | 99.28 (87.86) | **99.35** (88.30) |
| L-6 | 98.34 (82.93) | 97.76 (85.48) | 99.26 (87.24) | 98.85 (87.56) | **99.27** (87.79) | 99.25 (88.68) |
| L-7 | 98.10 (83.33) | 98.74 (84.34) | 98.93 (86.08) | **99.41** (87.22) | 99.32 (88.04) | 99.32 (88.21) |

TABLE VIII. LEARNING (OR CONVERGENCE) SPEED USING VARIOUS TRAINING EPOCHS. TABLE SHOWS INTER-SESSION GESTURE RECOGNITION ACCURACIES (%) ON TEST SET. THE NUMBERS ARE MAJORITY VOTED RESULTS USING 150 MS WINDOW (I.E., 150 FRAMES). PER-FRAME ACCURACIES ARE SHOWN IN PARENTHESIS.

| Model | # learning parameters | Training epochs | | | |
|---|---|---|---|---|---|
| | | 8 | 16 | 24 | 32 |
| Lightweight All-ConvNet+TL (Proposed) | $\approx 0.46\ M$ | **96.00** (71.56) | 96.60 (74.79) | 97.60 (76.92) | 97.69 (77.68) |
| Lightweight All-ConvNet-Slim (Proposed) | $\approx 0.19\ M$ | 91.92 (68.98) | 96.90 (73.70) | 98.28 (75.98) | **98.50** (77.47) |

*C. Gestures and the muscles involved in CapgMyo datasets*

Tables IX and X illustrate gestures and all the muscles involved in CapgMyo DB-a, DB-b and DB-c respectively [26].

TABLE IX. GESTURES IN CAPGMYO DB-A AND DB-B (8 ISOTONIC AND ISOMETRIC HAND CONFIGURATIONS)

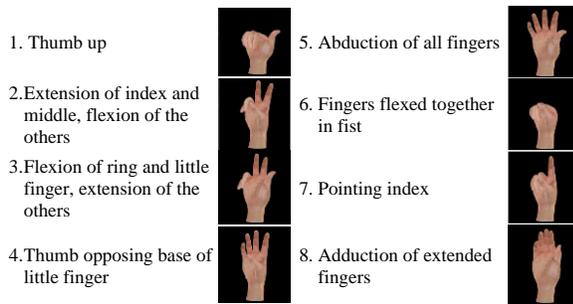

TABLE X. GESTURES IN CAPGMYO DB-C (12 BASIC MOVEMENTS OF THE FINGERS)

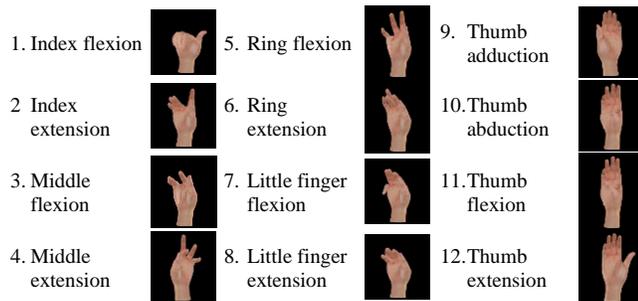